\documentclass[10pt,twocolumn,letterpaper]{article}

\usepackage{iccv}
\usepackage{times}
\usepackage{epsfig}
\usepackage{graphicx}
\usepackage{amsmath}
\usepackage{amssymb}
\usepackage{mathtools,nccmath}
\usepackage{physics}
\usepackage{multirow}
\usepackage{subfig}

\usepackage[pagebackref=true,breaklinks=true,letterpaper=true,colorlinks,bookmarks=false]{hyperref}

\iccvfinalcopy 

\def\iccvPaperID{8} 

\begin{document}

\title{A HVS-inspired Attention to Improve Loss Metrics for CNN-based Perception-Oriented Super-Resolution}

\author{Taimoor Tariq, Juan Luis Gonzalez and Munchurl Kim\\
Korea Advanced Institute of Science and Technology (KAIST)\\
{\tt\small $\{$taimoor.tariq, juanluisgb, mkimee$\}$@kaist.ac.kr}
\\
}

\maketitle


\begin{abstract}
   Deep Convolutional Neural Network (CNN) features have been demonstrated to be effective perceptual quality features. The perceptual loss, based on feature maps of pre-trained CNN's has proven to be remarkably effective for CNN based perceptual image restoration problems. In this work, taking inspiration from the the Human Visual System (HVS) and visual perception, we propose a spatial attention mechanism based on the dependency human contrast sensitivity on spatial frequency. We identify regions in input images, based on the underlying spatial frequency, which are not generally well reconstructed during Super-Resolution but are most important in terms of visual sensitivity. Based on this prior, we design a spatial attention map that is applied to feature maps in the perceptual loss and its variants, helping them to identify regions that are of more perceptual importance. The results demonstrate the our technique improves the ability of the perceptual loss and contextual loss to deliver more natural images in CNN based super-resolution. 
\end{abstract}
\section{Introduction}
    Inverse problems such as Single image super-resolution (SISR) and de-noising have been a subject of research for quite some time. Recently, Convolutional Neural Networks (CNNs) have been showed remarkable results for SISR. Traditionally, the emphasis of CNN based SISR has been distortion oriented, where the goal is to minimize the distortion using metrics such as the Peak Signal to Noise Ration (PSNR) and Structural Similarity Index (SSIM). Recently though, perception oriented super-resolution has gained wide interest, where the goal is deep learning techniques to get the best perceptual quality of restored images, maintaining naturalness and photo-realistic appearance. This paper addresses loss metrics for perception oriented super-resolution. 
    
    The perceptual loss proposed by Johnson et al. \cite{7} was one of the first to demonstrate how effective the deep CNN features can be as Full-Reference perceptual quality features, especially when used in loss functions for image restoration. The perceptual loss is now a benchmark and standard loss function popularly adopted in many image to image translation problems such as super-resolution, style transfer, denoising etc. \cite{9},\cite{10},\cite{11}. Zhang et al. \cite{12} and Blau et al. \cite{5} further demonstrate how effective the deep features really are as perceptual quality features. More recently, Mecherez et al. \cite{38} proposed a variation of the perceptual loss called the contextual loss, which still employs deep CNN features as perceptual quality features but uses an approximation of the KL-divergence to quantify distance. The contextual loss has been demonstrated to be quite effective in maintaining natural image statistics during SISR. The recent PIRM Super-Resolution Challenge Report \cite{29} clearly iterates that the perceptual loss and the contextual loss are the most widely used loss functions for CNN based perceptual image Super-Resolution.
    
    In this work, we exploit a known and well established property of the HVS called contrast sensitivity. Psychovisual experiments have revealed that the visual system's ability to respond to contrast, varies significantly with spatial frequency of the stimulus \cite{14}. Neuro-scientists have been able to quantify this variation using a function called the Contrast Sensitivity Function (CSF) \cite{32}. Considering that it is an important representative factor of our visual perception, the CSF has been used extensively by the community to improve image processing algorithms and techniques \cite{32}. 
    
    We use a variation of the CSF filtering technique of images to extract maps which represent regions which the human visual system is most sensitive to in terms of contrast. The map is then applied to feature maps of pre-trained networks used in the perceptual loss and contextual loss to give more emphasis to spatially important areas during restoration. The results will demonstrate that the technique improves the perceptual quality of restored images and delivers a better perception-distortion trade-off compared to the standard and popularly used perceptual loss \cite{7} and contextual loss \cite{38}.

\begin{figure} [t!]
\includegraphics[width=0.5\textwidth]{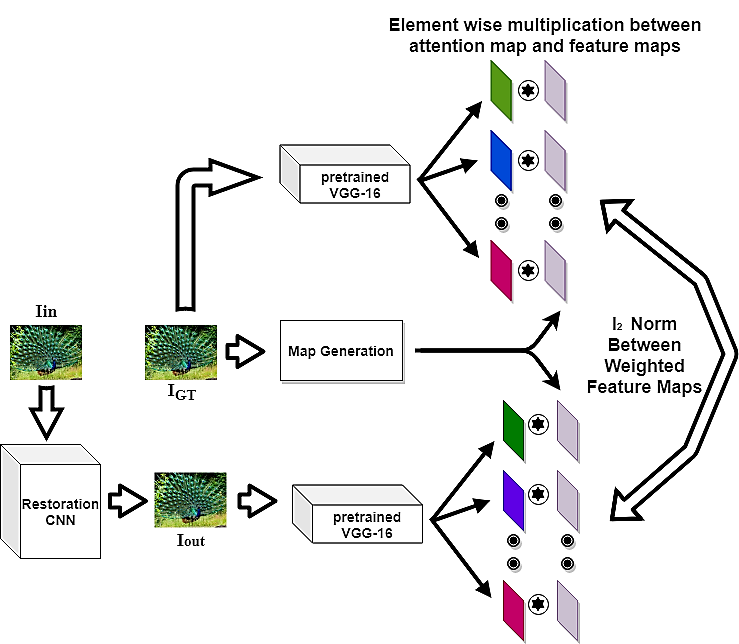}
\caption{The distance between feature representations of reference and restored images is generally used as a perceptual metric in the perceptual loss and its variants. We propose a spatial attention map, applied to feature representations to enable them to focus on more important regions during the learning process.}
\label{fig:Spike_Sorting}
\end{figure}

\section{A HVS inspired Attention}

\subsection{Motivation}
In SISR, the low resolution image ($I_{LR}$) can be modeled as a low-pass filtered and down-sampled version of the high-resolution image ($I_{HR}$). In the CNN based inverse problem of estimating $I_{HR}$ from $I_{LR}$, there can be an infinite number of of possible HRs that have the same low-frequency components (relatively easily reconstructed) but different high frequency components. Therefore, the goal for perception oriented SISR should put more emphasis on perceptually important frequency components. 

The net response/sensitivity of the HVS to different spatial frequencies is summarized in the CSF. Fig. 2(a) shows the human CSF derived through psychovisual experiments. The units of the CSF are in cycles/degree. This unit translates to the number of cycles per visual degree of the observer. Therefore, considering the SISR problem, we will design a CSF based spatial attention mechanism to help perceptual losses put more emphasis on reconstructing perceptually important spatial information, where there is minimal contrast masking in human perception.

\subsection{CSF based Map Generation}
Let $\mathit{I_{GT}}$ be the Y channel of the ground truth image with size MxN. The 2D DFT of the image will be
\begin{equation}
\mathit{F[u,v]}=\frac{1}{MN}\sum\limits_{m=0}^{M-1}\sum\limits_{n=0}^{N-1}I_{GT}[m,n].e^{-j2\pi(\frac{u}{M}m+\frac{v}{N}n)}
\end{equation}
Where 'u' and 'v' are the horizontal and vertical frequency indicators respectively.
As iterated earlier, the CSF is defined on the spatial frequency scale of cycles per visual degree. The current frequency units of the DFT need to be appropriately converted into the required scale for effective application of the CSF. Let, 
\begin{equation}
f(u) = \frac{u-1}{\Delta M} , f(v) = \frac{v-1}{\Delta N}
\end{equation}
where $\Delta$ = 0.25mm is the dot pitch. The cycles/degree equivalent of the DFT frequency units with a viewing distance of 'dis' millimeters will be $\mathit{s(u,v)}$ \cite{33} where, 

\begin{equation}
s(u,v) = \frac{\pi}{180\times\arcsin(\frac{1}{\sqrt{1+dis^2}})}\times\sqrt{f(u)^2+f(v)^2}
\end{equation}

Now we have effectively expressed the DFT on the spatial frequency units of cycles/degree. To generate a map that represents the most important spatial information, we will approximate the input image using a narrow band of spatial frequencies around the peak of the CSF. Let, 
\begin{equation}
 \hat{F}(u,v)=\begin{cases}
    F(u,v), & \text{$s_l$ $\leq$ s(u,v) $\leq$ $s_h$}.\\
    0, & \text{otherwise}.
  \end{cases}
\end{equation}
Now we take the inverse 2D DFT of the approximated representation to generate a map as shown in Eq. (5). 
\begin{equation}
\mathit{T[m,n]}=\sum\limits_{u=0}^{M-1}\sum\limits_{v=0}^{N-1}\hat F[u,v].e^{j2\pi(\frac{u}{M}m+\frac{v}{N}n)}
\end{equation}

As a final step, we normalize the matix 'T' so that its largest element is unity. $\|T\|_M$ is the max norm of T in Eq.(6) (largest element of the matrix). 
\begin{equation}
\mathit{\mu}=\frac{1}{\|T\|_M}\times T
\end{equation}

The viewing distance for HDTV displays should be selected for a 30 degree viewing angle. Considering parameters for viewing distance and for effective perception of common distortions (checkerboard artifacts etc) introduced by the use of the perceptual loss, we selected the viewing distance to be 55cm. We empirically selected the range of the filter to be the frequencies corresponding to 45\% of the peak of the CSF i.e $s_l$=2 cycles per degree and $s_u$=23 cycles per degree. 
\begin{figure}[hbt!]
\centering
\subfloat[CSF]{\includegraphics[width=0.5\textwidth]{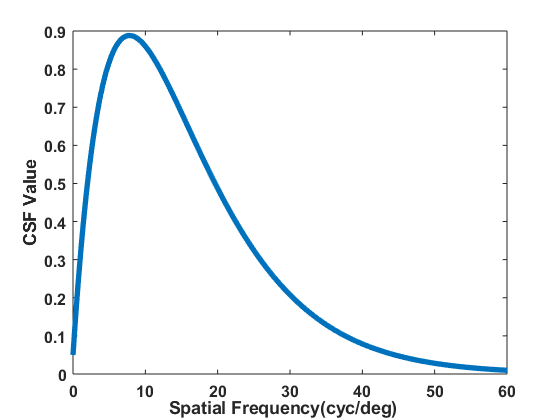}\label{fig:f5}}
\hfill
\subfloat[Input Image]{\includegraphics[width=0.23\textwidth]{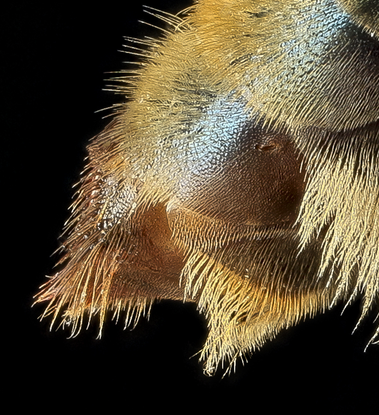}\label{fig:f7}}
\hfill
\subfloat[CSP based map]{\includegraphics[width=0.23\textwidth]{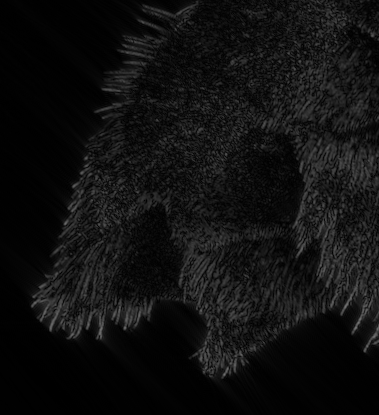}\label{fig:f7}}
\hfill
\caption{A Human Contrast Sensitivity Function (CSF), a sample input image and its corresponding attention map.}
\end{figure}

\section{Loss Functions for CNN based Perception-Oriented SISR}
\subsection{Perceptual Loss}
Deep CNN features are the result of non-linear transformations on images into a high dimensional manifold. The perceptual loss is based on the fact that distance between features on the transformed manifold might result in a better perceptual quality measure compared to distance between image pixels themselves. The high dimensional manifold in this case is the manifold of CNN features. The standard form of the perceptual loss, as per \cite{7} is given in Eq.(7)
\begin{equation}
\mathit{l_p}=\frac{1}{M\times W\times H}\sum\limits_{m=1}^M\Vert\Phi^k_m(I\textsubscript{out})-\Phi^k_m(I\textsubscript{GT})\Vert_2^2
\end{equation}
Where '$\Phi^k_m$' is the '$\textit{m}^{th}$' feature map in the '$\textit{k}^{th}$' layer with '$\textit{M}$' number of feature maps with dimensions '$\textit{H$\times$W}$' . This approach and its variants have proven to be remarkably effective as perceptual features in FR-IQA methods \cite{25}, image restoration \cite{10} and style transfer \cite{11} problems. The perceptual loss is a benchmark loss function in perceptual image restoration as demonstrated in \cite{29}. 
\subsection{Contextual Loss}
The contextual loss \cite{38} takes deep feature representations of images as input and aims to minimize an approximated version of the KL-Divergence. Let $x_i = \Phi^k_i(I\textsubscript{out})$ and $y_i = \Phi^k_i(I\textsubscript{GT})$ and the normalized pairwise distance $d_{ij}=dist(x_i,y_j)/(\min_k dist(x_i,y_j)+\epsilon)$. The pair-wise affinities are firstly defined as:
\begin{equation}
    A_{ij}=\frac{\exp(1-d_{ij}/h)}{\sum\limits_{l}\exp(1-d_{il}/h)}
\end{equation}
The contextual loss between feature distribution's of the $k-th$ layer of the pre-trained CNN (eg. VGG-16) can be expressed as
\begin{equation}
    l_{cx}(x^k,y^k)=-\log(\frac{1}{N}\sum\limits_{j}\max_i A_{ij})
\end{equation}
As demonstrated in \cite{38}, the contextual loss helps maintain natural image statistics in images during SISR. 
\subsection{Proposed Spatially Attentive Loss Function's}
The perceptual loss in the form of Eq. (7) does not spatially discriminate between information in feature maps on the basis of their perceptual importance. We propose extending the perceptual loss to a form in Eq. (10).
\begin{equation}
\mathit{l_p^{att}}=\frac{1}{M\times W\times H}\sum\limits_{m=1}^M\Vert \mu^k(I\textsubscript{GT})\odot(\Phi^k_m(I\textsubscript{out})-\Phi^k_m(I\textsubscript{GT}))\Vert_2^2    
\end{equation}
where $u^k(I\textsubscript{GT})$ is the CSF based attention map. The superscript 'k' represents that is has been appropriately re-sized to match the feature map dimensions in the k-th layer. The formulation in Eq. (10) will help the perceptual loss to give more importance in restoration of regions which are perceptually more important and have a visual susceptibility to perception of distortions which result in a loss of contrast, commonly associated with Super-Resolution and Blurring. 

For the contextual loss, similarly, instead of inputting the set of feature maps $\Phi^k(I\textsubscript{out})$ and $\Phi^k(I\textsubscript{GT})$, we will input the feature maps weighted by out attention map i.e $\mu^k(I\textsubscript{GT})\odot\Phi^k(I\textsubscript{out})$ and $\mu^k(I\textsubscript{GT})\odot\Phi^k(I\textsubscript{GT})$. The modified contextual loss will thus be:
\begin{equation}
    l_{cx}^{att}=l_{cx}(\mu^k(I\textsubscript{GT})\odot x^k,\mu^k(I\textsubscript{GT})\odot y^k)
\end{equation}

The results will demonstrate that our proposed extension improves both the perceptual and contextual loss, delivering a much better perception-distortion trade-off when used for CNN based SISR. 

\section{Experimental Setup}
\subsection{Overview}
To demonstrate the superiority of our proposed extension over the classical and widely used perceptual loss, we will make use of two distinct experimental techniques. The first being objective quality assessment (OQA) tests and the second being an image restoration experiment. 
\subsection{Objective Quality Test}
OQA tests are a standard method to evaluate the effectiveness of perceptual quality metrics. The best judges of perceptual quality are human observers. OQA tests try to correlate the performance of objective metrics with human judgment of perceptual quality. The more correlated a metric is with human subjective assessment of quality, the better it can be inferred as a perceptual quality metric. For human subjective assessment of perceptual quality, we will use the LIVE subjective image quality database \cite{24}. The data-set consists of human subjective scores in the form of Differential Mean Opinion Scores (DMOS) for a large number of images with a wide variety of distortions. The correlation of objective metric scores with human subjective scores is quantified using three metrics namely the Root Mean Square Error (RMSE), Linear Correlation Coefficient (LCC) and the Spearman Rank Order Correlation Coefficient (SROCC) after curve fitting of the objective and subjective scores using a non-linear polynomial. 

For comparing the objective metrics in Eq. -(7) and -(10), we will use feature maps from different layers of a wide variety of pre-trained networks such as the VGG-16 \cite{30}, ResNet-18 \cite{34}, SqueezeNet \cite{35} and AlexNet \cite{36} to reinforce the validity of our approach. We will also repeat our OQA test for distortions such as Gaussian Blur and Multiple Distortions \cite{37} (camera image acquisition process where images are first blurred due to narrow depth of field or other de-focus and then corrupted by white Gaussian noise to simulate sensor noise). 
\subsection{Super-Resolution Experiments}
Our second experiment will be an x4 Image Super-Resolution experiment using the well known VDSR network trained on the standard DIV2K data-set using the using the perceptual and contextual loss separately in  in combination with the $\mathit{l_2}$ loss. For the perceptual loss, we will train the network will network with;
\begin{equation}
L_{p} = \alpha.\mathit{l}\textsubscript{2} + (1-\alpha).\mathit{l}\textsubscript{p}
\end{equation}
and 
\begin{equation}
L_{p}^{att} = \alpha.\mathit{l}\textsubscript{2} + (1-\alpha).\mathit{l}\textsubscript{p}^{att}
\end{equation}
and for the contextual loss, we will train the network with
\begin{equation}
L_{cx} = \alpha.\mathit{l}\textsubscript{2} + (1-\alpha).\mathit{l}\textsubscript{cx}
\end{equation}
and 
\begin{equation}
L_{cx}^{att} = \alpha.\mathit{l}\textsubscript{2} + (1-\alpha).\mathit{l}\textsubscript{cx}^{att}
\end{equation}
\subsubsection{Perception-Distortion Trade-off}
To evaluate and compare the improved performance of our extensions in Eq.(13) and Eq.(15), we will compare the perception-distortion trade-off they deliver in comparison to the standards in Eq.(12) and Eq.(14) respectively. The Structural Similarity Index (SSIM) and the Peak Signal to Noise Ratio (PSNR) have traditionally, somewhat incorrectly been used as metrics to evaluate image restoration techniques. These metrics measure distortion between two images and are not sufficient to quantify the perceptual quality. Recent works such as \cite{5} and \cite{12} have pointed out that to quantify the efficacy of perception-oriented image restoration techniques, we need to analyze something known as the perception-distortion trade-off and perceptual quality. In recent works \cite{29}, The perceptual quality has been quantified using a metric known as perceptual index (PI) which is a combination of two No-Reference quality metrics called the Naturalness Image Quality Evaluater (NIQE) \cite{28} and the NRQM \cite{31}. The PI of an image $I_{in}$ is thus defined in Eq.(16)
\begin{equation}
PI(I_{in}) = \frac{1}{2}((10-NRQM(I_{in}))+NIQE(I_{in}))
\end{equation}
The lesser the PI, the better the perceptual quality of the image. The distortion (measured by the SSIM and PSNR) and the perceptual quality (PI) are in a trade-off relation, as explained in \cite{5}. The effectiveness of one technique over another can be quantified by its ability to deliver a better trade-off i.e lesser PI at the same SSIM. We will sufficiently demonstrate that our attention improves the perception-distortion trade-off for both the perceptual loss and contextual loss.
\begin{table*}[]
\centering
\caption{The attention module improves the correlation with human subjective scores of the modified perceptual loss ($l_p^{att}$) in Eq.(10) compared to the widely used standard perceptual loss ($l_p$) in Eq.(7) which leads to the conclusion that $l_p^{att}$ is a better objective perceptual quality metric. The results are repeated across multiple pre-trained networks and their layers.}
\begin{tabular}{|l|l|l|l|l|}
\hline
\multicolumn{5}{|l|}{\textbf{(SqueezeNet)   -   (Multiple Distortions)}}                                           \\ \hline
\textbf{Layer}                                   & \textbf{Metric} & \textbf{RMSE} & \textbf{LCC} & \textbf{SROCC} \\ \hline
\multirow{2}{*}{\textbf{Fire2-ReLU\_expand3x3}}  & $l_p$              & 15.2720       & 0.6132       & 0.5589         \\ \cline{2-5} 
                                                 & $l_p^{att}$            & 13.9830       & 0.6906       & 0.6084         \\ \hline
\multirow{2}{*}{\textbf{Fire4\_ReLU\_expand1x1}} & $l_p$              & 14.5749       & 0.6570       & 0.6051         \\ \cline{2-5} 
                                                 & $l_p^{att}$            & 12.9300       & 0.7434       & 0.6688         \\ \hline
\multirow{2}{*}{\textbf{Fire6\_ReLU\_expand3x3}} & $l_p$              & 15.5951       & 0.5910       & 0.5482         \\ \cline{2-5} 
                                                 & $l_p^{att}$            & 14.3458       & 0.6704       & 0.6135         \\ \hline
\end{tabular}
\end{table*}

\begin{table*}[]
\renewcommand\thetable{1.(b)}
\caption{}
\centering
\begin{tabular}{|l|l|l|l|l|}
\hline
\multicolumn{5}{|l|}{\textbf{(SqueezeNet)   -   (Gaussian Blur)}}                                                  \\ \hline
\textbf{Layer}                                   & \textbf{Metric} & \textbf{RMSE} & \textbf{LCC} & \textbf{SROCC} \\ \hline
\multirow{2}{*}{\textbf{Fire2-ReLU\_expand3x3}}  & $l_p$              & 11.0894       & 0.7185       & 0.7375         \\ \cline{2-5} 
                                                 & $l_p^{att}$            & 10.1010       & 0.7737       & 0.8010         \\ \hline
\multirow{2}{*}{\textbf{Fire3\_ReLU\_expand1x1}} & $l_p$              & 10.5569       & 0.7494       & 0.7641         \\ \cline{2-5} 
                                                 & $l_p^{att}$            & 8.2405        & 0.8561       & 0.8712         \\ \hline
\multirow{2}{*}{\textbf{Fire4\_ReLU\_expand1x1}} & $l_p$              & 10.2650       & 0.7652       & 0.7867         \\ \cline{2-5} 
                                                 & $l_p^{att}$            & 8.5886        & 0.8425       & 0.8591         \\ \hline
\end{tabular}
\end{table*}

\begin{table}[]
\renewcommand\thetable{1.(c)}
\caption{}
\begin{tabular}{|l|l|l|l|l|}
\hline
\multicolumn{5}{|l|}{\textbf{(VGG-16)   -   (Gaussian Blur)}}                                        \\ \hline
\textbf{Layer}                     & \textbf{Metric} & \textbf{RMSE} & \textbf{LCC} & \textbf{SROCC} \\ \hline
\multirow{2}{*}{\textbf{ReLU2\_2}} & $l_p$              & 10.3392       & 0.7612       & 0.7766         \\ \cline{2-5} 
                                   & $l_p^{att}$            & 8.5227        & 0.8451       & 0.8694         \\ \hline
\multirow{2}{*}{\textbf{ReLU3\_2}} & $l_p$              & 9.4553        & 0.8052       & 0.8288         \\ \cline{2-5} 
                                   & $l_p^{att}$            & 8.7198        & 0.8372       & 0.8572         \\ \hline
\multirow{2}{*}{\textbf{ReLU4\_2}} & $l_p$              & 9.4181        & 0.8069       & 0.8301         \\ \cline{2-5} 
                                   & $l_p^{att}$            & 8.7319        & 0.8367       & 0.8453         \\ \hline
\end{tabular}
\end{table}

\begin{table}[]
\renewcommand\thetable{1.(d)}
\caption{}
\begin{tabular}{|l|l|l|l|l|}
\hline
\multicolumn{5}{|l|}{\textbf{(VGG-16)   -   (JPEG2000)}}                                             \\ \hline
\textbf{Layer}                     & \textbf{Metric} & \textbf{RMSE} & \textbf{LCC} & \textbf{SROCC} \\ \hline
\multirow{2}{*}{\textbf{ReLU1\_1}} & $l_p$              & 7.4194        & 0.8917       & 0.8836         \\ \cline{2-5} 
                                   & $l_p^{att}$            & 5.4533        & 0.9431       & 0.9342         \\ \hline
\multirow{2}{*}{\textbf{ReLU2\_2}} & $l_p$              & 7.0756        & 0.8992       & 0.8669         \\ \cline{2-5} 
                                   & $l_p^{att}$            & 5.7946        & 0.9355       & 0.9285         \\ \hline
\multirow{2}{*}{\textbf{ReLU3\_2}} & $l_p$              & 6.7050        & 0.9126       & 0.9024         \\ \cline{2-5} 
                                   & $l_p^{att}$            & 6.1821        & 0.9262       & 0.9167         \\ \hline
\end{tabular}
\end{table}

\begin{table}[]
\renewcommand\thetable{1.(e)}
\caption{}
\begin{tabular}{|l|l|l|l|l|}
\hline
\multicolumn{5}{|l|}{\textbf{(AlexNet)   -   (Gaussian Blur)}}                                      \\ \hline
\textbf{Layer}                    & \textbf{Metric} & \textbf{RMSE} & \textbf{LCC} & \textbf{SROCC} \\ \hline
\multirow{2}{*}{\textbf{ReLU\_1}} & $l_p$              & 8.8626        & 0.8313       & 0.8450         \\ \cline{2-5} 
                                  & $l_p^{att}$            & 7.5581        & 0.8805       & 0.8914         \\ \hline
\multirow{2}{*}{\textbf{Conv\_2}} & $l_p$              & 9.3736        & 0.8089       & 0.8281         \\ \cline{2-5} 
                                  & $l_p^{att}$            & 6.4939        & 0.9133       & 0.9216         \\ \hline
\multirow{2}{*}{\textbf{ReLU\_2}} & $l_p$              & 9.5549        & 0.8006       & 0.8229         \\ \cline{2-5} 
                                  & $l_p^{att}$            & 0.8294        & 0.8327       & 0.8532         \\ \hline
\end{tabular}
\end{table}

\begin{table}[]
\renewcommand\thetable{1.(f)}
\caption{}
\begin{tabular}{|l|l|l|l|l|}
\hline
\multicolumn{5}{|l|}{\textbf{(ResNet-18)   -   (Multiple Distortions)}}                                 \\ \hline
\textbf{Layer}                        & \textbf{Metric} & \textbf{RMSE} & \textbf{LCC} & \textbf{SROCC} \\ \hline
\multirow{2}{*}{\textbf{Res2a\_ReLU}} & $l_p$              & 13.8896       & 0.6956       & 0.6522         \\ \cline{2-5} 
                                      & $l_p^{att}$            & 12.9627       & 0.7419       & 0.6717         \\ \hline
\multirow{2}{*}{\textbf{Res3a\_ReLU}} & $l_p$              & 11.1586       & 0.8166       & 0.7805         \\ \cline{2-5} 
                                      & $l_p^{att}$            & 9.4828        & 0.8714       & 0.8273         \\ \hline
\multirow{2}{*}{\textbf{Res4a\_ReLU}} & $l_p$              & 9.4211        & 0.8732       & 0.8448         \\ \cline{2-5} 
                                      & $l_p^{att}$            & 9.0407        & 0.8839       & 0.8518         \\ \hline
\end{tabular}
\end{table}

\begin{figure}[hbt!]
\centering
\subfloat[$l_p$]{\includegraphics[width=0.238\textwidth]{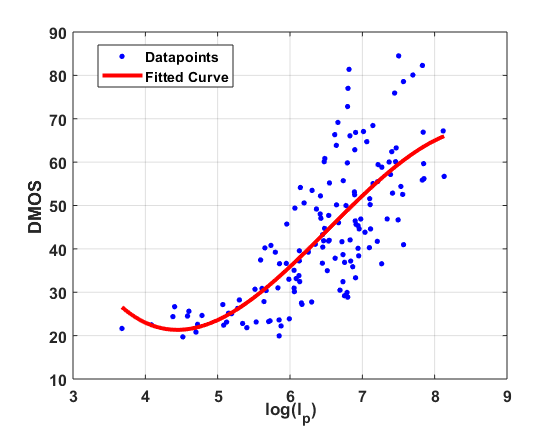}\label{fig:f5}}
\hfill
\subfloat[$l_p^{att}$]{\includegraphics[width=0.238\textwidth]{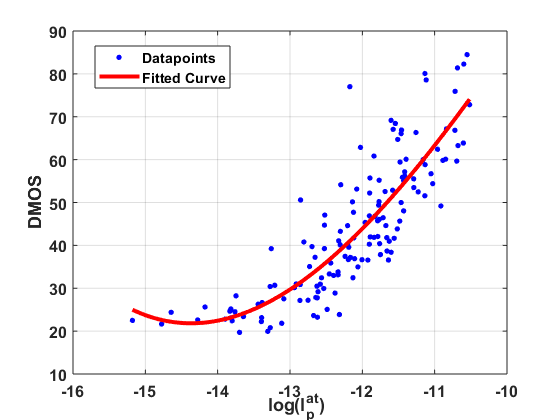}\label{fig:f7}}
\hfill
\caption{It can be seen that our proposed extension ($l_p^{att}$) is more correlated with human subjective DMOS compared to the perceptual loss ($l_p$). This result is for loss with feature maps '\textit{Fire3\_ReLU\_expand1x1}' layer of the SqueezeNet on Gaussian Blur distorted images.}
\end{figure}

\begin{figure} [t!]
\includegraphics[width=0.5\textwidth]{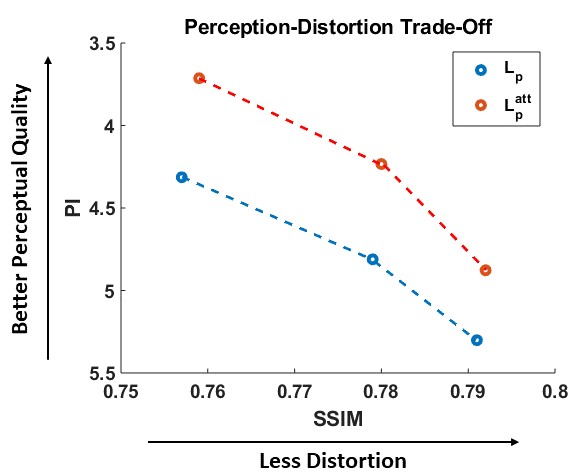}
\caption{The proposed attentive perceptual loss ($L_p^{att})$ significantly improves the perception-distortion trade-off compared to the widely used perceptual loss ($L_p$) in an x4 SR experiment with the VDSR on the DIV2K data-set. The three different points in each curve are obtained by varying the values of $\alpha$ in Eq.-(12) and -(13).}
\label{fig:Spike_Sorting}
\end{figure}

\subsection{Results and Discussions}
\subsubsection{Objective Quality Assessment Experiments}
Table. 1  demonstrates the correlation of the perceptual loss (Eq. 7) and our proposed extension (Eq. 10) with human subjective assessment of perceptual quality. The experiment is repeated over multiple pre-trained networks and their layers. It can be observed that our proposed extension is much more correlated with human subjective assessment of perceptual quality and thus a superior metric. The superiority is quantified in higher LCC and SROCC scores. A graphical representation is shown in Fig. 3 where it can be seen that the objective metric in Fig. 3-(b) is more correlated with human subjective DMOS compared to the one in Fig.3-(a). The next results will demonstrate that this effectiveness enables our proposed extension to improve perceptual image restoration compared to the traditional and widely used perceptual loss. 
\subsubsection{x4 Super Resolution Experiments}
For our x4 SR experiment, we trained a VDSR on the DIV2K data-set on both the loss functions in Eq.-(12) and -(13) under the exact same training conditions, using the '\textit{ReLU3\_2}' layer of the VGG-16 for the loss functions. We performed three training and testing trials for each loss function by varying the control parameter $\alpha$. The higher the $\alpha$ in Eq.-(12) and -(13), the more weight is given to the pixel-wise $l_2$ loss compared to the $l_p$ or $l_p^{att}$ loss causing an increase in the SSIM and a decrease in the perceptual quality (higher PI). This behavior is in accordance with the perception-distortion trade-off \cite{5}. Fig. 4 shows that even if the $\alpha$ parameter is varied, our proposed attentive perceptual loss always delivers a significantly better perception-distortion trade-off compared to the widely used perceptual loss \cite{7}. At roughly the same average SSIM on the test set, training with $L_p^{att}$ gives delivers images with much better perceptual quality. Furthermore, even at much lower distortions, the perceptual loss fails to achieve as good perceptual quality as our proposed extension delivers at much higher distortions, indicating a very significant improvement.  

Fig. 4 shows some crops from the DIV2K test set images restored using the VDSR trained on Eq.-(12) and -(13). It can be seen that at the same level of distortions (SSIM), the attentive perceptual loss ($L_p^{att}$) improves the perceptual quality (lower PI) compared to the perceptual loss ($L_p$). Close examination of the crops also reveals that the $L_p^{att}$ significantly helps in suppressing checkerboard artifacts, which are a common nuisance while using the perceptual loss. For example, while comparing Fig.5-(c) and -(d), the artifacts are clearly visible in Fig.5-(d) whereas Fig.5-(c) has a much smoother all round texture. Similar observations can be made in Fig.5-(h) and -(i) if the blue ocean background is carefully observed. 

Similarly, for an x4 SR experiment on the VDSR, Fig. 6 demonstrates the effectiveness of our attention in improving the ability of the contextual loss to deliver more natural images during CNN based SISR. Training the VDSR with Eq. (15) delivers a much better perception-distortion trade-off compared to training it with Eq. (14).  This results clearly demonstrate the efficacy of our proposed attention. As the perceptual loss and contextual loss are the most widely used loss functions in perceptual Super-Resolution, as iterated in \cite{29}, the proposed technique has high practical impact in CNN based Super-Resolution.

\begin{figure*}[hbt!]
  \centering
  \subfloat[Ground Truth]{\includegraphics[width=0.3333\textwidth]{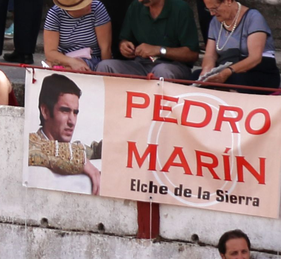}\label{fig:f5}}
  \hfill
  \subfloat[{$L_p^{att}$} (SSIM: 0.744, PI: 3.3497)]{\includegraphics[width=0.333334\textwidth]{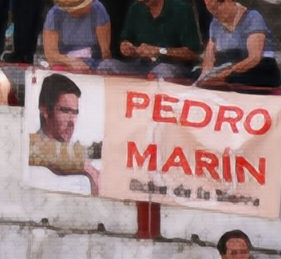}\label{fig:f7}}
  \hfill
  \subfloat[{$L_p$} (SSIM: 0.744, PI: 3.9229)]{\includegraphics[width=0.333\textwidth]{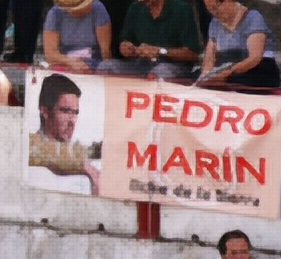}\label{fig:f6}}
  \hfill
  \subfloat[Ground Truth]{\includegraphics[width=0.333\textwidth]{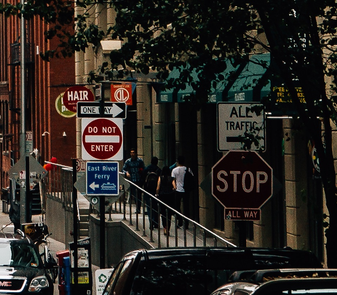}\label{fig:f5}}
  \hfill
  \subfloat[{$L_p^{att}$} (SSIM: 0.680, PI: 3.4226)]{\includegraphics[width=0.333\textwidth]{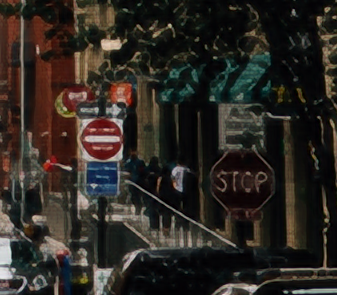}\label{fig:f7}}
  \hfill
  \subfloat[{$L_p$} (SSIM: 0.680, PI: 3.7100)]{\includegraphics[width=0.333\textwidth]{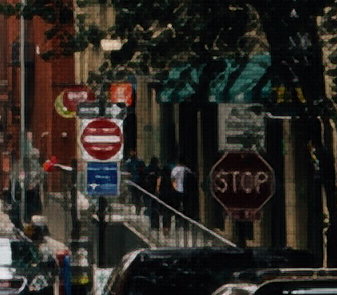}\label{fig:f6}}
  \hfill
  \subfloat[Ground Truth]{\includegraphics[width=0.333\textwidth]{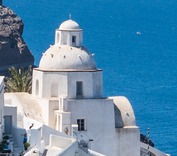}\label{fig:f5}}
  \hfill
  \subfloat[{$L_p^{att}$} (SSIM: 0.777, PI: 3.2619)]{\includegraphics[width=0.333\textwidth]{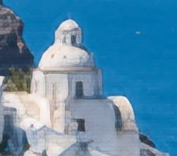}\label{fig:f7}}
  \hfill
  \subfloat[{$L_p$} (SSIM: 0.778, PI: 4.0202)]{\includegraphics[width=0.333\textwidth]{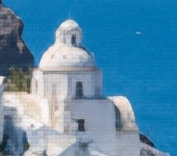}\label{fig:f6}}
  \hfill
  \caption{Single Image x4 SR on the VDSR \cite{26} architecture. Our proposed attentive perceptual loss ($L_p^{att}$) in Eq.(13) delivers better perceptual quality(less PI) at same levels of distortion(SSIM) compared to the perceptual loss ($L_p$) in Eq.(12). }
\end{figure*}
 
 \begin{figure*} [t!]
 \centering
\includegraphics[width=0.5\textwidth]{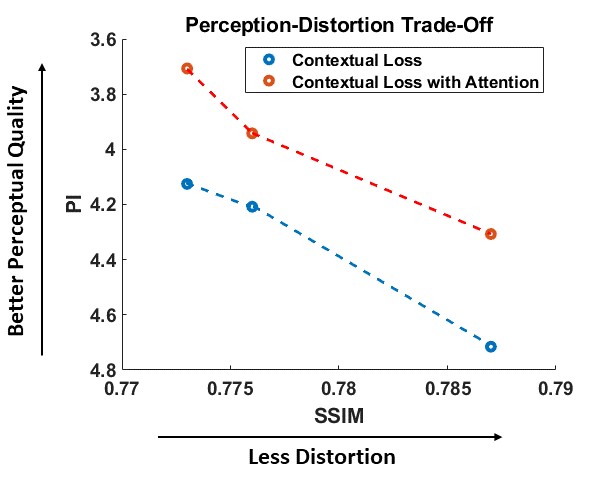}
\caption{The proposed attentive contextual loss ($L_{cx}^{att})$ significantly improves the perception-distortion trade-off compared to the contextual loss ($L_{cx}$) in an x4 SR experiment with the VDSR on the DIV2K data-set. The three different points in each curve are obtained by varying the values of $\alpha$ in Eq.-(14) and -(15).}
\label{fig:Spike_Sorting}
\end{figure*}  
 
\begin{figure*}[h!]
  \centering
  \subfloat[Ground Truth]{\includegraphics[width=0.3333\textwidth]{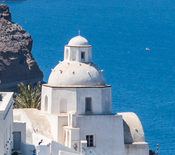}\label{fig:f5}}
  \hfill
  \subfloat[{$L_{cx}^{att}$} (SSIM: 0.786, PI: 3.9865)]{\includegraphics[width=0.333334\textwidth]{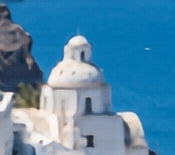}\label{fig:f7}}
  \hfill
  \subfloat[{$L_{cx}$} (SSIM: 0.786, PI: 4.5587)]{\includegraphics[width=0.333\textwidth]{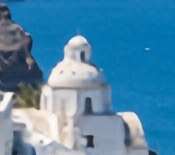}\label{fig:f6}}
  \hfill
  \subfloat[Ground Truth]{\includegraphics[width=0.333\textwidth]{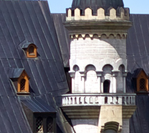}\label{fig:f5}}
  \hfill
  \subfloat[{$L_{cx}^{att}$} (SSIM: 0.742, PI: 4.0813)]{\includegraphics[width=0.333\textwidth]{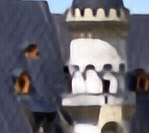}\label{fig:f7}}
  \hfill
  \subfloat[{$L_{cx}$} (SSIM: 0.741, PI: 4.7239)]{\includegraphics[width=0.333\textwidth]{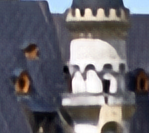}\label{fig:f6}}
  \hfill
  \caption{x4 SISR on the VDSR \cite{26} architecture. Our proposed attentive contextual loss ($L_{cx}^{att}$) in Eq.(15) delivers better perceptual quality (less PI) at same levels of distortion (SSIM) compared to the contextual loss ($L_{cx}$) in Eq.(14). }
\end{figure*}

\section{Conclusions}
Taking inspiration from the Human Visual System (HVS), more specifically the spatial frequency dependence of contrast sensitivity in our visual perception, we propose an attention mechanism to improve the widely used perceptual loss and contextual loss for perception-oriented SISR. Through objective quality assessment experiments, we verify that our proposed attentive perceptual loss is a better perceptual quality metric compared to the perceptual loss, owing to a better correlation with human subjective assessment of quality. Through an x4 Super-Resolution experiment, we also demonstrate that our proposed attention has the ability significantly improve the perceptual loss and contextual loss to deliver more natural images. Considering that the perceptual loss and contextual loss are the two most widely used loss functions in perceptual image restoration, our proposed extensions can effectively be used to train novel image restoration CNNs for perception oriented Super-Resolution.

\bibliographystyle{ieee}

\end{document}